\documentclass[lettersize,journal]{IEEEtran}
\usepackage{cite}

\usepackage{amsmath,amsfonts}
\usepackage{algorithmic}
\usepackage{algorithm}
\usepackage{array}
\usepackage[caption=false,font=normalsize,labelfont=sf,textfont=sf]{subfig}
\usepackage{textcomp}
\usepackage{stfloats}
\usepackage{url}
\usepackage{verbatim}
\usepackage{graphicx}

\usepackage{xcolor}
\usepackage{siunitx}
\usepackage{mathtools}
\usepackage{gensymb}
\usepackage{amssymb}  
\usepackage{soul} 
\newtheorem{theorem}{Theorem}
\newtheorem{remark}{Remark}
\newtheorem{assum}{Assumption}
\newtheorem{property}{Property}

\DeclareMathOperator{\diag}{diag}

\begin{document}

\title{An Integrated Approach to Aerial Grasping: Combining a Bistable Gripper with Adaptive Control}

\author{Rishabh Dev Yadav$^{1}$, Brycen Jones$^{2}$, Saksham Gupta$^{1}$, Amitabh Sharma$^{1}$,\\ Jiefeng Sun$^{2}$, Jianguo Zhao$^{2}$, and Spandan Roy$^{1}$
\thanks{This work was funded by a joint project on `Aerial manipulation with bistable gripper' supported by the I-HUB Foundation for Cobotics (IHFC) in India and the National Science Foundation (NSF) in the U.S. The first two authors contributed equally. Corresponding authors: J. Zhao and S. Roy.}
\thanks{$^{1}$Rishabh Dev Yadav, Saksham Gupta, Amitabh Sharma and Spandan Roy are with the Robotics Research Centre, International Institute of Information Technology Hyderabad, Hyderabad 500032, India
(e-mail: \{rishabh.yadav, saksham.g, amitabh.sharma\}@research.iiit.ac.in, spandan.roy@iiit.ac.in).}
}

\maketitle

\begin{abstract}
Grasping using an aerial robot can have many applications ranging from infrastructure inspection and maintenance to precise agriculture. 
However, aerial grasping is a challenging problem since the robot has to maintain an accurate position and orientation relative to the grasping object, while negotiating various forms of uncertainties (e.g., contact force from the object). 
To address such challenges, in this paper, we integrate a novel passive gripper design and advanced adaptive control methods to enable robust aerial grasping. 
The gripper is enabled by a pre-stressed band with two stable states (a flat shape and a curled shape). 
In this case, it can automatically initiate the grasping process upon contact with an object. 
The gripper also features a cable-driven system by a single DC motor to open the gripper without using cumbersome pneumatics.  
Since the gripper is passively triggered and initially has a straight shape, it can function without precisely aligning the gripper with the object (within an $80$ mm tolerance).
Our adaptive control scheme eliminates the need for any a priori knowledge (nominal or upper bounds) of uncertainties.   
The closed-loop stability of the system is analyzed via Lyapunov-based method.
Combining the gripper and the adaptive control, we conduct comparative real-time experimental results to demonstrate the effectiveness of the proposed integrated system for grasping.
Our integrated approach can pave the way to enhance aerial grasping for different applications.
\end{abstract}

\begin{IEEEkeywords}
Aerial Grasping, Aerial Robots, Adaptive Control, Mechanism Design, Aerial Manipulator.
\end{IEEEkeywords}

\section{Introduction}

Aerial grasping or manipulation involves the use of flying robots to physically interact with objects in the environment, enabling them to pick up, hold, move, or perform specific tasks with those objects. 
It offers significant advantages over traditional ground-based robots in various applications such as payload delivery in hazardous locations, inspection and maintenance of infrastructures like bridges, wind turbines, or power lines, etc. \cite{loianno2018localization, zhong2019practical, tognon2019truly, suarez2020benchmarks}. 
Overall, the development of aerial grasping and manipulation technologies in flying robots promises to revolutionize multiple industries by providing efficient, versatile, and safer alternatives to traditional methods.

\begin{figure}
    \centering
   \includegraphics[width=.5 \textwidth]{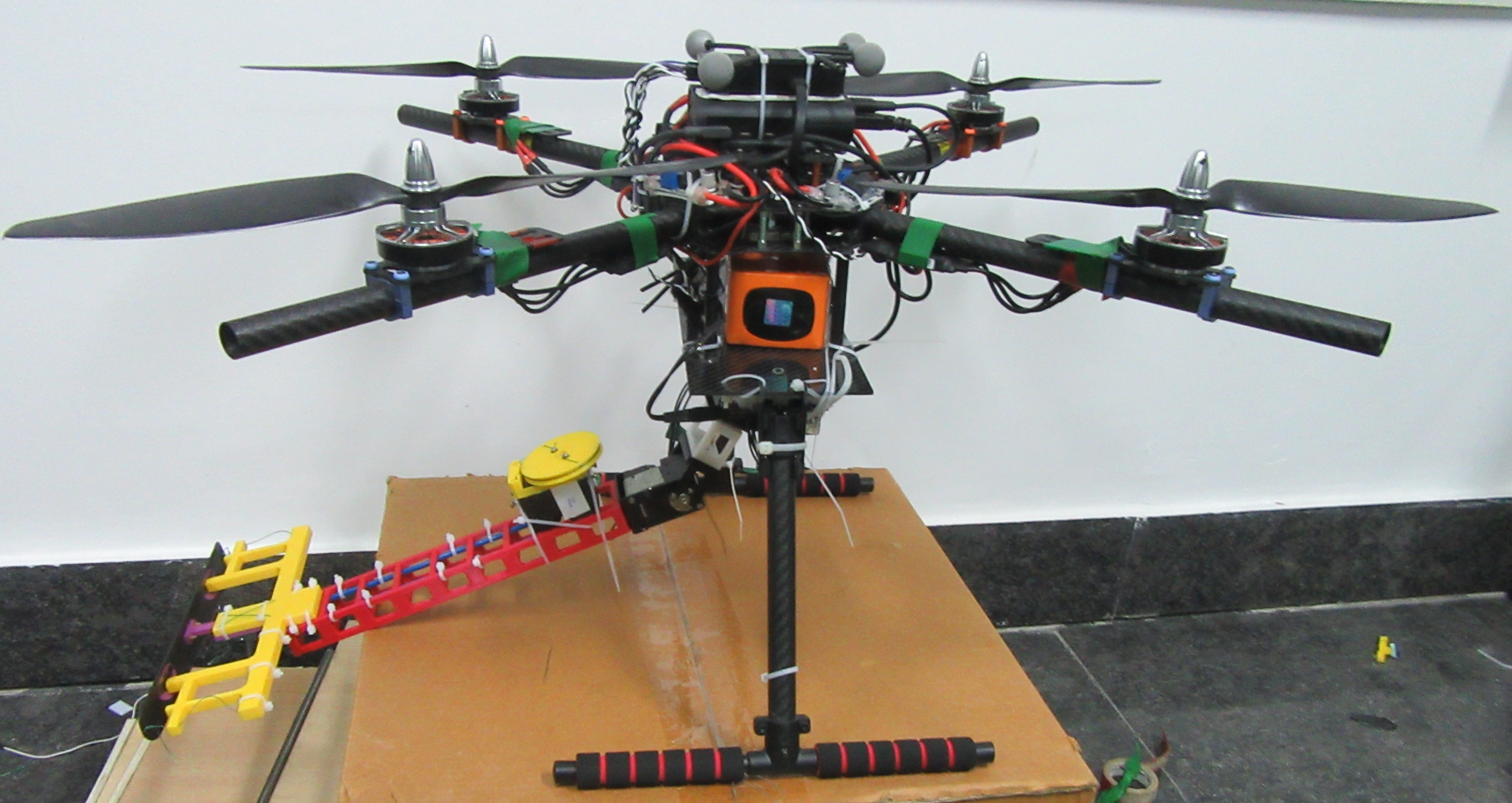}
    \caption{The developed aerial manipulation system includes the quadrotor, the two-link arm, and the bistable gripper.}
    \label{fig:whole_system}
\end{figure}

Despite the wide applications, performing robust aerial grasping is challenging.
To carry out aerial grasping, a combined system called Unmanned Aerial Manipulator (UAM) is typically used, where a quadrotor or multirotor carries a manipulator with an end-effector (cf. Fig. \ref{fig:whole_system}). 
To successfully grasp an object, the UAM has to maintain an accurate position and orientation relative to the grasping object, while negotiating various forms of uncertainties (e.g., payload variation, dynamic variation in center-of-mass of the system owing to manipulation, environmental disturbances, etc.). Furthermore, the grasping process can easily be disrupted by reaction forces resulting from the interaction between the gripper and the object. 
Therefore, developing successful aerial grasping demands a delicate balance between mechanical design for the gripper and controller design for the whole system. 

\subsection{Related Works}
In this section, we briefly review the relevant works from both gripper design and controller design aspects for aerial grasping or manipulation, discussing their drawbacks and highlighting the need for the proposed approach in this work.

Successful grasping of an object requires a proper design of the end-effector attached to the manipulator, which is generally realized as a gripper.
Some works have utilized traditional motor-driven gripper, which requires actuation to both close and open the gripper \cite{meng2022aerial}. 
Recently, several research groups have explored the use of passive grippers to grasp objects upon contact such as a spring-loaded claw \cite{stewart2022swoop}, a bistable claw with spring-connected fingers \cite{zufferey2022ornithopters}, 
a bistable gripper based on the von Mises bistable structure \cite{zhang2020compliant, hsiao2022mechanically}, 
a passive gripper with pre-stored potential energy \cite{chen2022aerial,roderick2021bird}. Nevertheless, most of these designs generally utilize a finger-like shape for grasping, resulting in relatively complicated designs.
Therefore, researchers have explored the use of pre-stressed bistable beams for grasping applications \cite{jitosho2019exploiting, wang2020soft, nguyen2022soft}.
However, all these designs use pneumatic actuation to open the gripper, resulting in a substantial weight and modification of pre-stressed beams (e.g., enclosing the beams with pneumatics).

For the control aspect, the main challenge arises from two sources.
First, the nonlinear coupling forces between the flying robot and manipulator, affect the behavior for both the components \cite{orsag2017dexterous}. 
Second, the inertial parameters change (e.g.,  center-of-mass and mass distribution) as the manipulator arm moves and interacts with the environment during grasping, object pick-and-drop, etc \cite{ruggiero2018aerial}. 
Based on the findings in review articles  \cite{ruggiero2018aerial, ollero2021past}, it has been observed that 
methods that require accurate knowledge of the system model (cf. \cite{suarez2018physical, kim2018cooperative,zhang2019robust, lee2020aerial, bicego2020nonlinear}), often fall short of the necessary control performance.

Accordingly, to tackle parametric uncertainties and external disturbances in UAM systems, researchers have used either disturbance observers \cite{kim2017robust, zhang2019robust, chen2020robust,lee2022rise, liang2021low, chen2022adaptive} or adaptive control \cite{liang2022adaptive}. 
However, methods using the disturbance observer require a priori bounds of uncertainties and ignore state-dependent uncertainties, which are inherent in practical electro-mechanical systems, such as those stemming from inertial parametric uncertainties \cite{roy2021adaptive}. Though the adaptive control scheme \cite{liang2022adaptive} incorporates state-dependent uncertainty, it necessitates precise knowledge of the inertia matrix. 
Overall, 
the complexity of a UAM system makes it extremely challenging, if at all possible, to acquire a priori precise knowledge, including nominal and/or upper bounds, of uncertain dynamics terms (as required in \cite{kim2017robust, zhang2019robust, chen2020robust,lee2022rise, liang2021low, chen2022adaptive, liang2022adaptive}).

\subsection{Contributions}
To the best of our knowledge, the state-of-the-art UAM has yet to integrate 1) a passive gripper without using pneumatics and does not require precise alignment between the gripper and the object, and 2) an adaptive control solution capable of addressing unknown state-dependent forces. 
Toward this direction, we present an integrated solution for aerial grasping by combining a novel gripper design and advanced adaptive control for a quadrotor-based UAM system (cf. Fig. \ref{fig:whole_system}) which features the following contributions:
\begin{itemize}

\item Our gripper can automatically initiate the grasping process upon contact with the object and does not require modifications to commercially available pre-stressed beams. 
It features a cable-driven system by a single DC motor to open the gripper without using cumbersome pneumatics.  
Since the gripper is passively triggered and initially has a straight shape, it can function without precisely aligning the gripper with the object (within an $80$ mm tolerance).

\item Our control scheme eliminates the need for any a priori knowledge (nominal or upper bounds) of uncertainties (unlike \cite{kim2017robust, zhang2019robust, chen2020robust,lee2022rise, liang2021low, chen2022adaptive}) or of inertial dynamic terms (unlike \cite{liang2022adaptive}).   \end{itemize}

The closed-loop stability of the system is analyzed via Lyapunov-based method, and extensive comparative real-time experimental results demonstrate the effectiveness of the proposed integrated scheme over the state-of-the-art.

The rest of this paper is organized as follows: the design and characterization of the proposed bistable gripper are detailed in Section \ref{sec:gripper}. 
Section \ref{sec:formulation} describes the control problem, the proposed adaptive control solution, and its analysis.
Section 
\ref{sec:experiments} presents the experimental results for aerial grasping.
Finally, Section V concludes the work.

The following notations are used in this paper:  $|| (\cdot)||$ and $\lambda_{\min}(\cdot)$ denote 2-norm and minimum eigenvalue of $(\cdot)$, respectively; $I$ denotes identity matrix with appropriate dimension and $\diag\lbrace \cdot, \cdots, \cdot \rbrace$ denotes a diagonal matrix with diagonal elements $\lbrace \cdot, \cdots, \cdot \rbrace$.

\section{Gripper Design and Characterization}\label{sec:gripper} 

In this section, we detail the mechanical design of the whole system, especially the customized bistable gripper. 

\subsection{System introduction}
We have developed an Unmanned Aerial Manipulator (UAM) system that integrates adaptive control algorithms and a bistable gripper to enable robust aerial grasping. 
Our UAM system comprises three main components: 1) a commercially available S-650 quadrotor system that serves as the aerial platform, 2) a two-link planar manipulator with two revolute joints that provide the necessary dexterity for aerial manipulation tasks, and 3) a customized gripper that employs a pre-stressed bistable beam and a cable-driven system for object grasping.

Using this UAM system, a typical working process is that the quadrotor flies to a location close to an object, then the arm moves the gripper to contact the object.  
Upon contact, the bistable gripper can quickly grasp the object. 
The proposed control strategy (detailed in section \ref{sec:controller}) allows the aerial manipulation system to be stable while grasping. 
The two-link arm can change the gripper's angle and position to facilitate the grasping process. 

\subsection{Gripper Design}
\begin{figure} [!h]
    \includegraphics[width=.5 \textwidth]{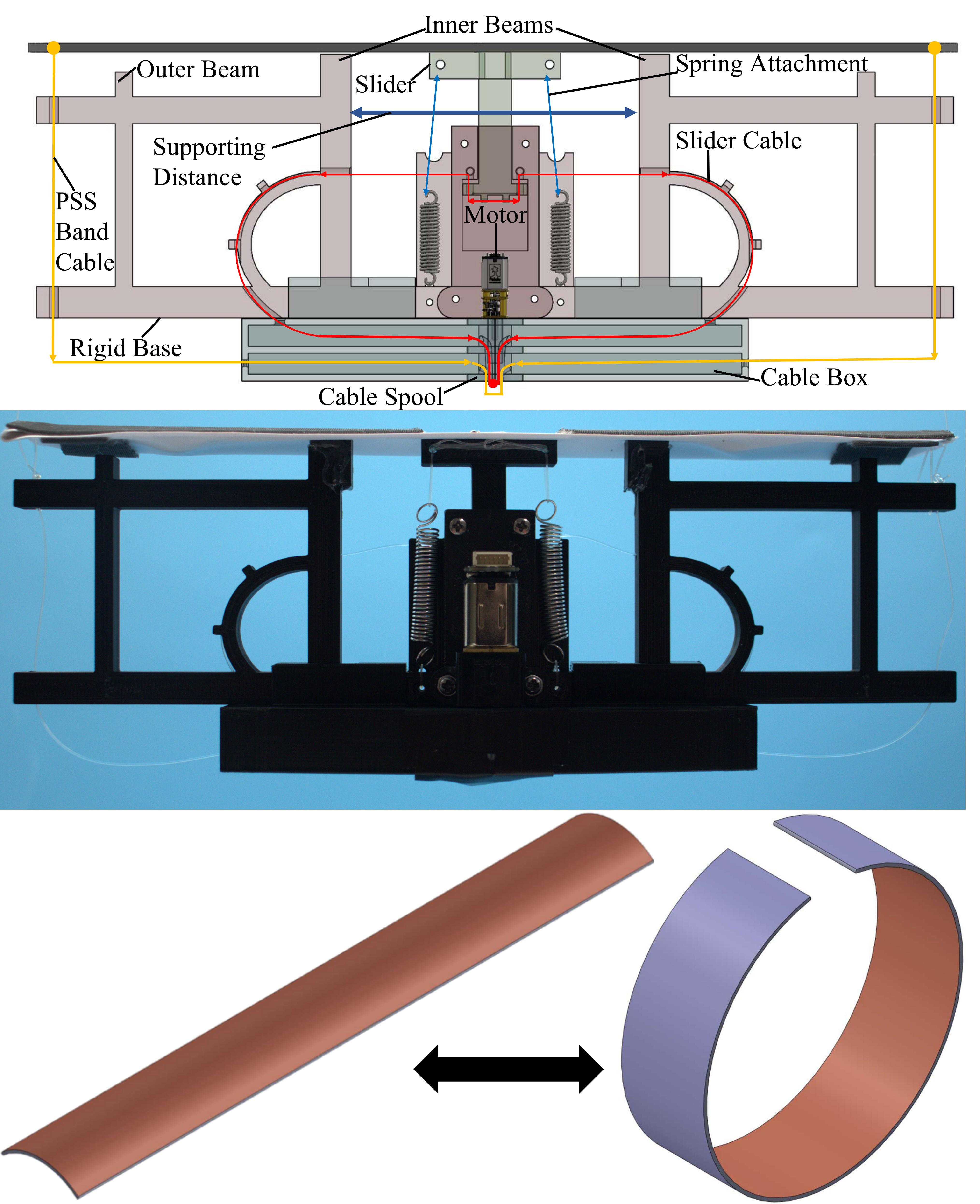}
    \centering
    \caption{The design of the bistable gripper. Top: a detailed 3D model with major components labeled; middle: the prototype of the gripper; bottom: illustration of the two stable states of the PSSB placed on top of the gripper.}
    \label{fig:gripper_design}
\end{figure}

Since the gripper is customized, we explain the design rationale in this sub-section.
The gripper mainly contains three major components: a $210$ mm x $27.5$ mm x $0.5$ mm pre-stressed spring steel band (PSSB) as fingers for grasping, a 3D-printed rigid base to hold the PSSB and other components, and a cable-driven system for opening the gripper (Fig. \ref{fig:gripper_design}).
The PSSB is initially straight but can curl to grasp an object (bottom of Fig. \ref{fig:gripper_design}). 
The middle of the PSSB is connected to the base through a slider that slides linearly up and down to allow the PSSB to retract during grasping.
The two sides of the PSSB are supported by two inner beams from the rigid base. 
If the PSSB is compressed by an impact force from an object between the two inner beams, the slider moves downward, causing the PSSB to curl rapidly once the force is greater than the triggering force $F_{tr}$. 
The PSSB can initiate the grasping process without additional actuators.
Because the PSSB is flexible, it can adapt to the object's shape after it is curled to grasp. 

Opening of the PSSB is realized through a cable-driven system actuated by a DC motor.
Two fishing lines (illustrated as yellow and red in Fig. \ref{fig:gripper_design}) with equal slack are spooled on a dual-cable spool attached to a $1000:1$ geared DC motor with a magnetic encoder (\#2373, Pololu).  
The inner loop (red) will push the slider upward, while the outer loop (yellow) will flatten the PSSB. 
The cables can be unspooled and loosened for reuse.
The inner cable loop is routed through a curved section to smoothly transit from the sliding mechanism to the spool. 
Both cable loops have approximately the same slack such that the flattening of the PSSB coincides with the extension of the sliding mechanism. 
Both cables are routed to be inside a cable management box to ensure the cables will not be tangled during the release process.

The rigid base is designed for optimal cable routing, such that the PSSB is flattened with the smallest force possible by pulling the PSSB open with a vertical downward force (the yellow cables are routed downward through holes in the base as seen in Fig.~\ref{fig:gripper_design}). 
There are two outer beams at the two further ends of the base to facilitate the flattening of the PSSB because the PSSB should be bent past a straight shape to return and maintain the flattened shape. Therefore, we design the two outer beams to be $4$ mm lower than the inner beams. 

Two linear springs (50465A, $7/32"$x$1"$, 14.55 mm, NEIKO) are used to apply pretension to PSSB to reduce the triggering force $F_{tr}$. 
The springs are attached from the base to the slider by fishing lines. 
The pretension can be manually adjusted by altering the displacement of the springs when the PSSB is flattened. 
The pretension in the springs can be estimated using $ F_{Spring} = k \Delta {x}$, where $k=0.274\pm.009$ N/mm is experimentally obtained by measuring force and displacement three times using a force gauge (M5-2, Mark-10) and calipers. 

\subsection{Gripper Characterization}

\begin{figure}[!b]
    \centering \includegraphics[width=.5\textwidth]{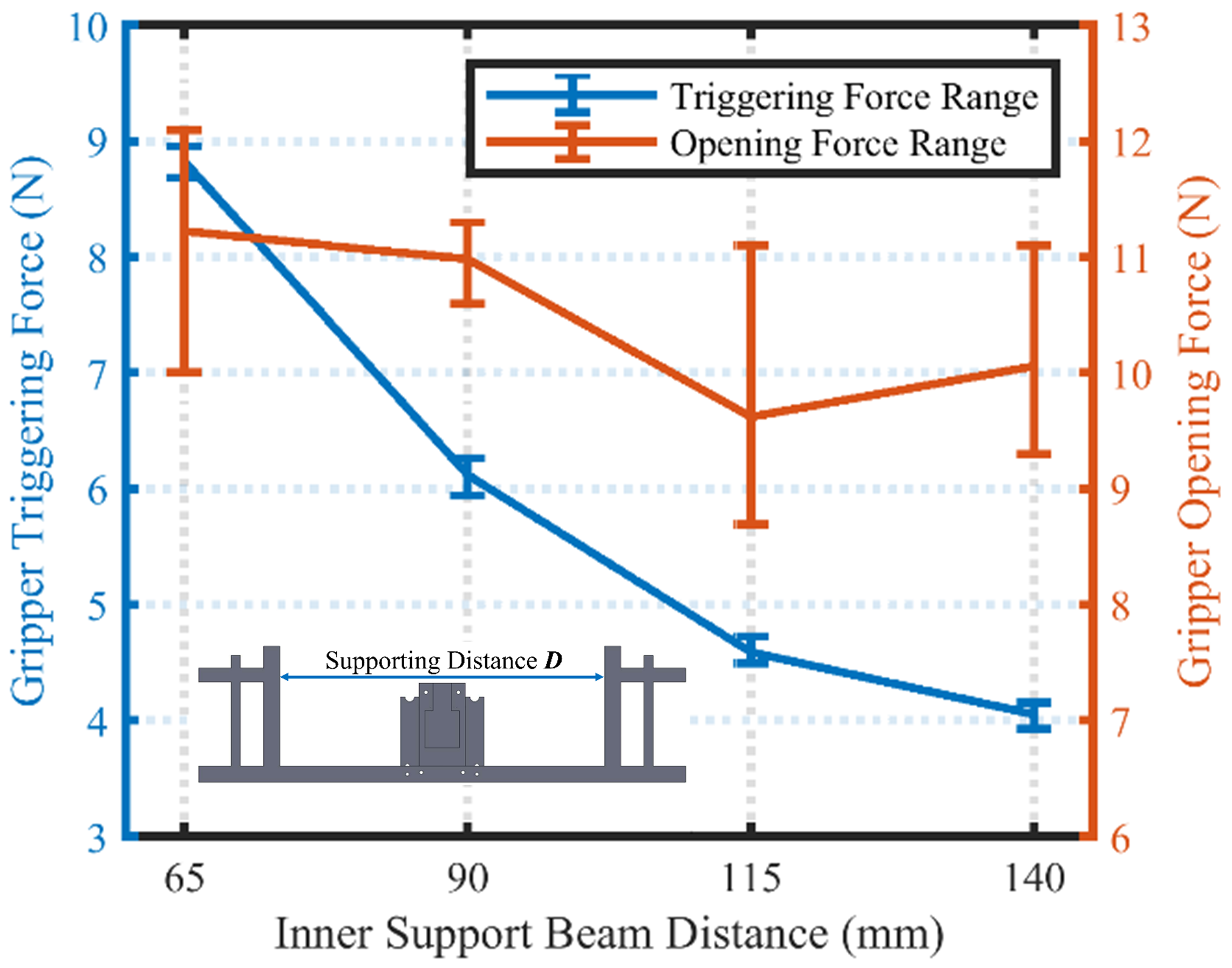 }    
    \caption{Triggering and opening force when  the support distance between the two inner beams varies from $65$ mm to $140$ mm.}
    \label{fig:OpenTrigForceBeamDist}
\end{figure}

With the designed gripper, this section will detail the characterization of the gripper's triggering and opening force because these two forces will determine how we should control the manipulator arm so that the gripper can successfully grasp an object.  
For the triggering force, a small triggering force is desired for robust gripping and dynamic aerial grasping. 
Therefore, we will evaluate the influence of various parameters such as the distance between the two inner beams, spring pretension force, and offset distance from the center. 
We also quantify the activation time since a fast activation is also desired because it will reduce the chances of a failed grasp of an object.

First, we examine how the support distance between the two inner beams will influence the triggering and opening force. 
The two inner beams support the PSSB  as the sliding mechanism is pushed downward (Fig. \ref{fig:gripper_design}), meaning the distance between them directly affects the triggering force. 
Four different rigid bases are 3D printed with inner beam distances of $65$, $90$, $115$, and $140$ mm.   
A force gauge (M5-2, Mark-10) is used to measure the triggering force with a test stand (ESM303, Mark-10). 
A 3D-printed base holds the gripper in the center of the test stand's lower jaw, and an $8$ cm diameter cylinder is centered over the gripper in the top jaws. 
The gripper and cylinder are aligned so off-axis forces do not affect the results. 
The bottom of the cylinder is set $5$ mm above the PSSB and lowered at a rate of $40$ mm/min until the gripper is activated and the force levels off. 
The gripper is then reset, and the experiment is repeated five times for each support distance. 
The results of this experiment are plotted in Fig. \ref{fig:OpenTrigForceBeamDist}, showing that the lowest triggering force is achieved at a $140$ mm support distance. 
Figure \ref{fig:OpenTrigForceBeamDist} shows the triggering force can be approximated by a second-order polynomial in terms of support distance: $y=0.008x^2 -0.24x +20.66$. 
This allows for the gripper design to be altered to achieve a desired triggering force before any pretension is applied. 

\begin{figure*}[!t]
    \centering
    \includegraphics[width=0.9\textwidth]{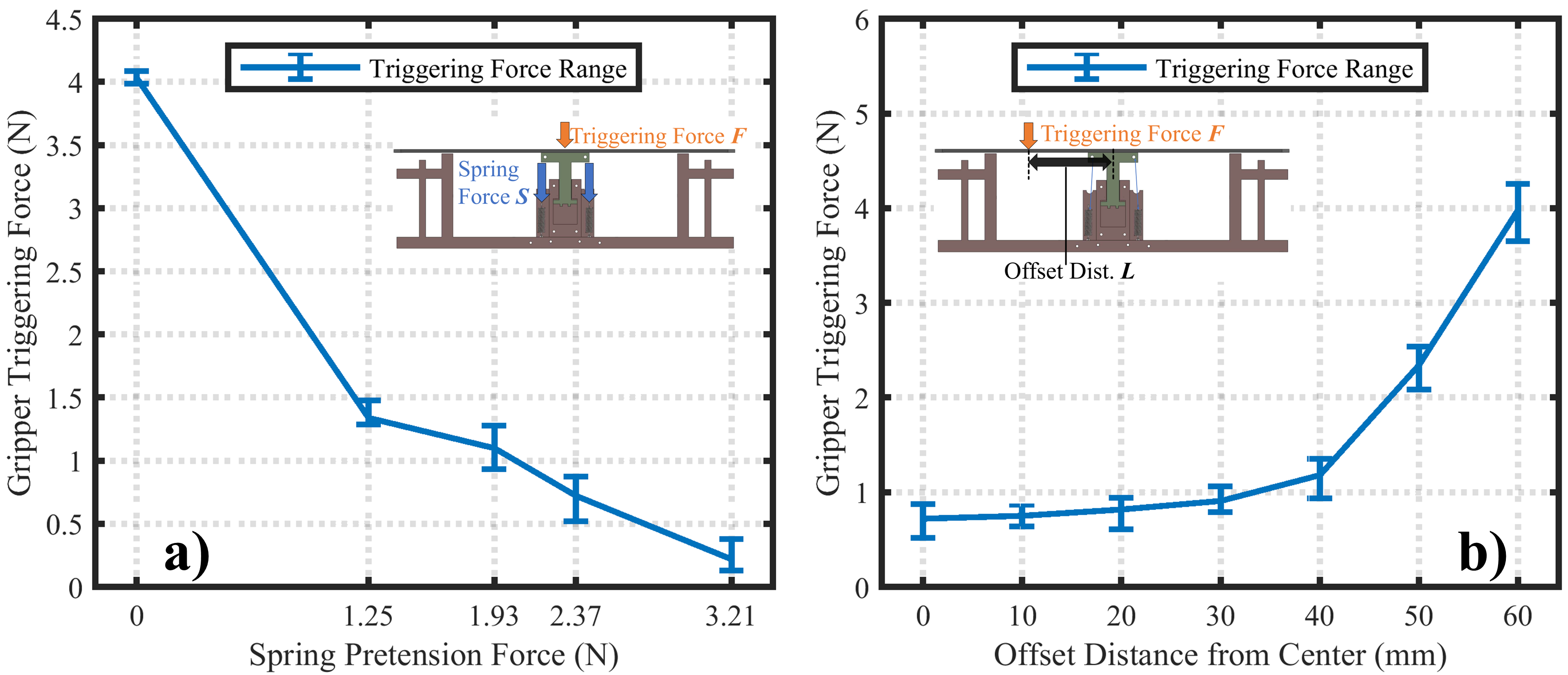}
    \caption{a) Triggering force with respect to four different spring pretensions ($1.25$ to $3.21$ N) for $140$ mm support distance. b) Triggering Force with respect to different offset distances ($0$ to $60$ mm) from the center for $140$ mm base with $2.37$ N of pretension.}
    \label{fig:PretenVTrig140beam}
\end{figure*}

To determine the effect of support distance on opening force, we test the opening force for the four rigid bases with no pretension force using another force gauge with a larger range (M2-20, Mark-10). The opening force is the force required to flatten both sides of the PSSB. 
Each rigid base is placed in a vice, with a fishing line running from either side of the PSSB through a pulley placed in the top jaws of the test stand to ensure equal force distribution to both sides of the band. 
The top jaw is raised at a speed of $330$ mm/min until both sides of the PSSB are fully flattened, which is indicated by a distinct `snap' sound. 
The experiment is repeated five times for each of the four bases.
We record the maximum force during each experiment as the opening force and plot the results in Fig. \ref{fig:OpenTrigForceBeamDist}, where the opening force decreases slightly between support distances of $65$ and $115$ mm before increasing slightly at $140$ mm. 
Since the opening force does not vary greatly with the support distance, we do not need to consider this to choose a good support distance. 
Based on the results of triggering and opening forces, we choose the base with a support distance of $140$ mm, as it provides the lowest triggering force before pretension is applied.
Note that a slightly larger supporting distance might be possible, but it will not significantly decrease the triggering force.

The smallest triggering force we can obtain without pretension is $4.05\pm0.11$ N with a support distance of $140$ mm. 
Since this value is still relatively large, we use linear springs to apply pretension to the PSSB.
To determine the relationship between the pretension applied through the springs and the triggering force of the PSSB, five pretension forces values of $0$, $1.25$, $1.93$, $2.37$, and $3.21$ N are applied.
Note that these forces are not evenly distributed since we can only estimate the pretension force by manually changing the length of the string that connects the spring to the slider. 
We conduct experiments for the triggering force under the five different pretension forces.
The results are shown in Fig.~\ref{fig:PretenVTrig140beam}, exhibiting a steep drop in triggering force between no pretension and $1.25$ N and a more gradual decrease between $1.25$ and $3.21$ N of pretension. 
From this experiment, we choose an ideal pretension range of $2$ to $2.5$ N, as this range allows for a  triggering force in the range of $0.5$ to $1$ N, which is small but stable (i.e., the gripper would not be triggered by external disturbance). 
Therefore, a pretension of $2.37$ N  is used for the remaining experiments.

For all the above characterizations, the triggering force is applied at the center of PSSB. 
The advantage of the developed gripper is that we do not need to precisely align the centers of the gripper and the object. 
In other words, the object can contact the PSSB at different locations, but successful grasping can still be ensured.
This is important because the aerial grasping process is dynamic where the contact point is unlikely to be perfectly centered on the gripper.
To assess the effect of contact locations on the triggering force, we test the triggering force again using the same methodology as above, with the contact location of a $8$ cm cylinder being offset from the center of PSSB with an offset distance from $10$ to $60$ mm (step size: $10$ mm). 
The results in Fig. \ref{fig:PretenVTrig140beam} show that, at less than $40$ mm, the offset distance has a small effect on the triggering force and, at larger than $40$ mm, it has a large effect (the triggering force almost quadruples for a distance of $60$ mm). 
This result suggests the gripper will still function and trigger as desired when the contact point is within an $80$ mm range along the PSSB. 

Finally, a quick gripper activation is desired for the best results when performing dynamic aerial grasping. 
We test the activation time using an $8$ cm cylinder at four different locations: centered, $15$ mm, $25$ mm, and $35$ mm offset from the center of the PSSB.  
\begin{figure}
    \includegraphics[width=.5\textwidth]{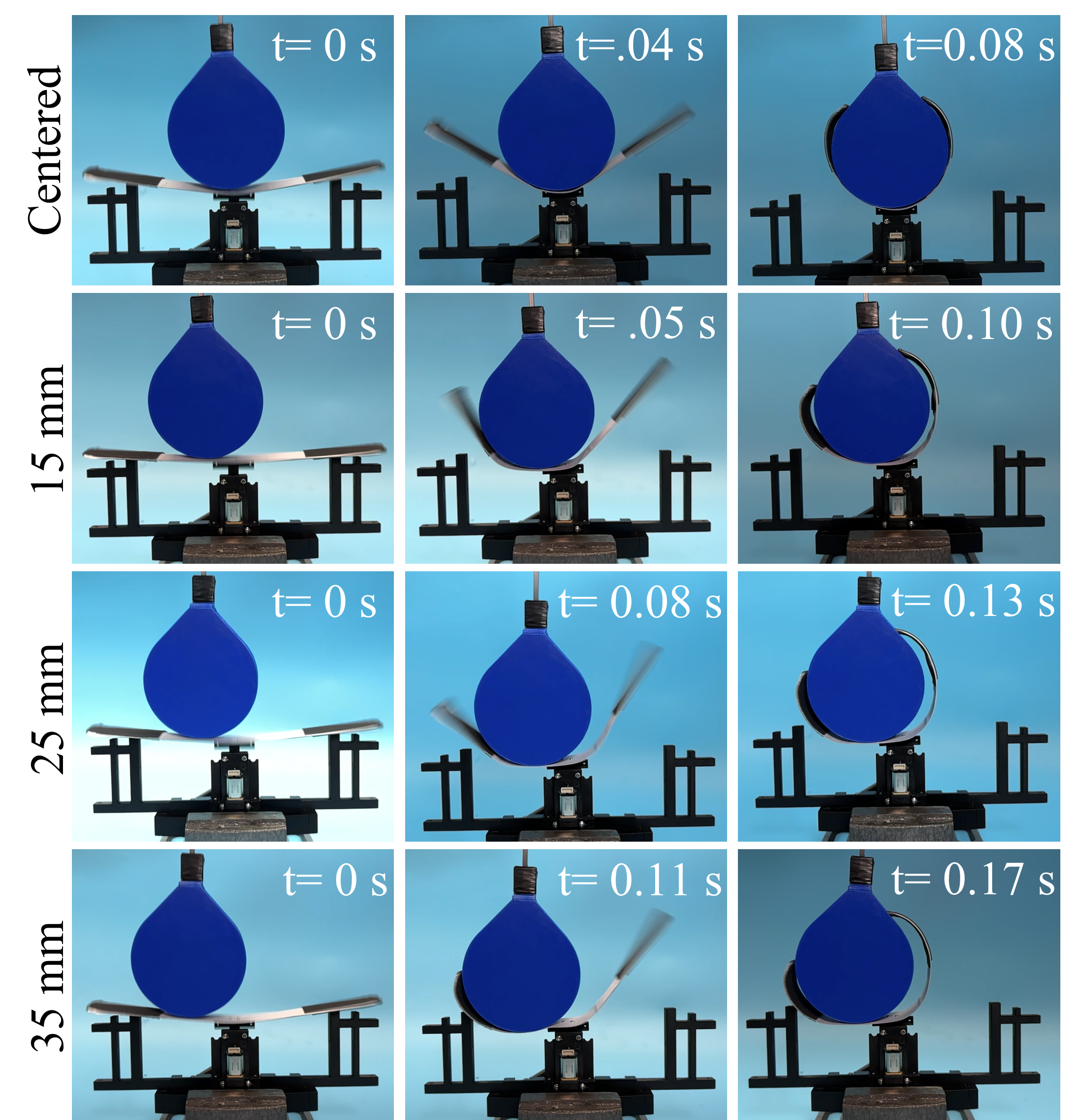}
    \centering
    \caption{Activation time for gripper with an $8$ cm cylinder with $0$, $15$, $25$, and $35$ mm offset from the center.}
    \label{fig:ActivTime}
\end{figure}
The cylinder is placed on the PSSB and is pressed down until the gripper is triggered. The activation process is filmed using a high-speed camera at $240$ frames per second (fps). 
The time $t=0$ is set on the first frame when the PSSB can be seen moving and is stopped on the first frame when the PSSB has stopped curling. 
The results of this experiment provide an activation time accurate to $\pm{.005}$ seconds. 
Each test is repeated five times, and the average times during activation are represented in Fig. \ref{fig:ActivTime}.
The results in Fig. \ref{fig:ActivTime} suggest that the activation time increases with respect to the offset distance with the longest activation time being $0.17$ $\pm{.005}$ s.
This is because one side of the PSSB needs more time to curl around the object when the offset distance is larger.
However, the maximum activation time is still within the ability of the adaptive control system to position the drone in the correct proximity to the object being gripped during the entire activation process.

\section{System Dynamics, Controller Design and Analysis} \label{sec:formulation} 

With the designed gripper, we can establish the system dynamics for the UAM and formulate an adaptive controller that can ensure object grasping while maintaining system stability under various unknown uncertainties.

\begin{figure}[!t]
\begin{center}
    \includegraphics[scale=0.4]{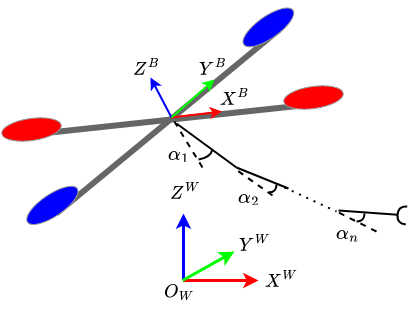}
    \caption{A schematic for a typical UAM system with an n-link manipulator and the corresponding frames.}
    \label{UAM_model}
\end{center}
\end{figure}

\begin{table}[!t]
\renewcommand{\arraystretch}{1.0}
\caption{{Nomenclature}}
\label{table 3}
\centering
{
{	\begin{tabular}{c c}
		\hline \\
	$[ X^B ~ Y^B ~Z^B ]$& Quadrotor body-fixed coordinate frame\\
	$[ X^W ~ Y^W ~ Z^W]$ & Earth-fixed coordinate frame \\
	$[x ~ y ~z]$ & Quadrotor position in $[ X^W ~ Y^W ~ Z^W]$\\
	$[\phi ~ \theta ~ \psi]$ & Quadrotor roll, pitch and yaw angles\\
	$[\alpha_1, ~ \alpha_2, ~\cdot \cdot ~, \alpha_n]$ & Manipulator joint angles\\
	$M\in \mathbb{R}^{(6+n)\times (6+n)}$ & Mass matrix \\ 
	$C\in \mathbb{R}^{(6+n)\times (6+n)}$ &  Coriolis matrix \\ 
	$g\in \mathbb{R}^{6+n}$&  Gravity vector \\
	$d\in \mathbb{R}^{6+n} $&  Bounded external disturbance \\
	$\tau_p, \tau_q \in \mathbb{R}^3$ & Generalized quadrotor control inputs \\
 $\tau_{\alpha} \in \mathbb{R}^n$ & Manipulator's joint control inputs \\
\hline 
\end{tabular}}}
\end{table}

The Euler-Lagrangian dynamical model of an aerial-manipulator system with an $n$ degrees-of-freedom (DoFs) planar manipulator system (cf. Fig. \ref{UAM_model}) is typically represented as \cite{arleo2013control,kim2013aerial}
\begin{subequations}
\begin{align}  
M(\chi)\ddot{\chi} + C(\chi, \dot{\chi})\dot{\chi} + g(\chi) + d &= [ \tau_p^T~ \tau_q^T ~ \tau_{\alpha}^T ]^T \label{eq:EL_dynamics}, \\
	   {\tau_{p}} &= {R^W_B U}, \label{con}
\end{align}
\end{subequations}
where the term $d$ represents the combination of \emph{bounded but unknown external forces (e.g., impact force between the manipulator-gripper and object) and disturbances (e.g., wind)}; $\chi \triangleq \begin{bmatrix}
x(t),y(t),z(t) ,\phi(t) , \theta(t) , \psi(t) ,
\alpha_1(t) , .., \alpha_n(t)
\end{bmatrix} $ and the other various symbols are explained in Table \ref{table 3}; $\tau_q (t)  \triangleq 
[u_2(t), ~ u_3(t),~ u_4(t)]$ is the control inputs for roll, pitch, and yaw of the quadrotor;  ${\tau_{p}} = {R^W_B U}$ is the generalized control input for quadrotor position in Earth-fixed frame, such that ${U}(t)\triangleq
\begin{bmatrix}
0 & 0 & u_1(t)
\end{bmatrix}^T$ with $u_1$ being the thrust force in the body-fixed frame and ${R_B^W} $ being the $Z-Y-X$ Euler angle rotation matrix describing the rotation from the body-fixed coordinate frame to the Earth-fixed frame \cite{mellinger2011minimum}
\begin{align}
	{R_B^W} =
	\begin{bmatrix}
	c_{\psi}c_{\theta} & c_{\psi}s_{\theta}s_{\phi} - s_{\psi}c_{\phi} & c_{\psi}s_{\theta}c_{\phi} + s_{\psi}s_{\phi} \\
	s_{\psi}c_{\theta} & s_{\psi}s_{\theta}s_{\phi} + c_{\psi}c_{\phi} & s_{\psi}s_{\theta}c_{\phi} - c_{\psi}s_{\phi} \\
	-s_{\eta} & s_{\phi}c_{\theta} & c_{\theta}c_{\phi}
	\end{bmatrix}, \label{rot_matrix}
	\end{align}
	where $c_{(\cdot)}$, $s_{(\cdot)}$ are abbreviations for $\cos{(\cdot)}$,  $\sin{(\cdot)}$, respectively.

The following standard system properties for the UAM hold owing to the Euler-Lagrange mechanics \cite{spong2006robot, kim2013aerial}:
\begin{property} \label{prop_1}
The matrix {$M(\chi)$} is uniformly positive definite and $ \exists \underline{m},\overline{m} \in \mathbb{R}^{+}$ such that $0 < \underline{m}I \leq M(\chi) \leq \overline{m}I$. 
\end{property}

\begin{property} \label{prop_2}
$\exists \bar{c}, \bar{g}, \bar{d} \in\mathbb{R}^{+}$ such that $||C (\chi)|| \leq \bar{c}||\dot{\chi}||$, $||g (\chi)|| \leq \bar{g}$ and $||d|| \leq \bar{d}$. 
\end{property}

\begin{property} \label{prop_3}
The matrix $(\dot{{M}} - 2{C})$ is skew symmetric, i.e., for any non-zero vector $r$, we have $r^T(\dot{{M}} - 2{N})r = 0$.
\end{property}

\subsection{Control Objective, Challenges, and Problem Formulation} For a successful grasp, the control system must ensure the followings: 
\begin{itemize}
    \item Positioning of the drone and the gripper within a triggering proximity of the object to apply the required triggering force. This requires following some desired trajectory for the drone and the manipulator, simultaneously. 
    \item Tackling the reaction force stemming from the interaction between the gripper and objects (subsumed under $d$ in Eq.~(\ref{eq:EL_dynamics})). Such a force is difficult to model precisely, as it depends on the objects to be grasped and the positioning of the gripper as per application requirements (cf. the experimental scenarios). Left unattended, this force can critically compromise the control performance (as it propagates throughout the system), leading to an unsuccessful grasp. 
    \item Achieving the above two objectives also requires efficient handling of the inevitable parametric uncertainty (due to unknown payload) and external disturbances (e.g., wind). In addition, manipulation 
    activity with a payload creates dynamic variation in center-of-mass which orchestrates significant uncertainty in the inertial parameters.   
\end{itemize}
In view of the above discussions, we represent the amount of system parametric uncertainties in the form of the following assumption, which, in fact, acts as a control design challenge. 

\begin{assum} [Uncertainty]
The system dynamics terms $M, C, g, d$ and their bounds $\overline{m}, \underline{m}, \bar{c}, \bar{g}, \bar{d}$ defined in Properties 1-2 are unknown for control design.
\end{assum}

Let $\chi_d (t)$ be a desired trajectory to be followed, where $\chi_d (t)$ and its time-derivatives $\dot{\chi}_d(t), \ddot{\chi}_d(t)$ are 
designed to be smooth and bounded. The control problem is defined as:

\textit{Control Problem:} Under system Properties 1-3, design an adaptive control law to track a desired trajectory $\chi_d (t)$ without any a priori knowledge of system parameters and external disturbances (cf. Assumption 1).

\subsection{Proposed Adaptive Controller Design} \label{sec:controller}
The aforementioned control problem is solved in this sub-section. We first define an error variable $s$ as
\begin{equation}
    s (t) \triangleq \dot{e} (t) + \Phi e (t),
\label{slidingVar}
\end{equation}
where $ e = \chi - \chi_d $ and $\Phi$ is a positive definite gain matrix. In the following, we shall remove the variable dependencies for brevity.

Multiplying the
time derivative of (\ref{slidingVar}) by ${M}$ and using (\ref{eq:EL_dynamics}) yields
\begin{equation}
    {M}\dot{s} =  {M}(\ddot{\chi} - \ddot{\chi}_d + \Phi \dot{e}) = \tau - {C}s + \varphi
\label{dot_s}
\end{equation}
where $\tau=[ \tau_p^T~ \tau_q^T ~ \tau_{\alpha}^T ]^T$; $\varphi \triangleq  -(C \dot{\chi} + g +  d + M \ddot{\chi}_d  - M \Phi \dot{e} - Cs )$ represents the \emph{overall system uncertainty}, and its upper bound structure using Properties 1-2 is computed as 
\begin{align}
    ||\varphi||  \leq & \bar{d} + \bar{g} + \bar{c}||\dot{\chi}||^2 + \bar{m}(||\ddot{\chi_d}|| + ||\Phi||||\dot{e}||) \nonumber \\
    & + \bar{c}||\dot{\chi}||(||\dot{e}|| + ||\Phi||||\chi||).
\label{boundness}
\end{align}
Defining $\xi \triangleq [e^T~ \dot{e}^T]^T$ , using the inequalities $||\xi|| \geq ||e||$ and $||\xi|| \geq ||\dot{e}||$, and substituting $\dot{\chi} = \dot{e} + \dot{\chi_d}$ in (\ref{boundness}) yields
\begin{equation}
    ||\varphi|| \leq  {K}^*_0 + {K}^*_1 ||\xi|| + {K}^*_2 ||\xi||^2
\label{up_bound}
\end{equation}
where ${K}^*_0 = \bar{d} + \bar{g} + \bar{c} ||\dot{\chi}_d||^2 + \bar{m}||\ddot{\chi_d}||$, ${K}^*_1 =  \bar{c} ||\dot{\chi}_d||(3 + ||\Phi||) + \bar{m}||\Phi||$ and
${K}^*_2 = \bar{c}(2 + ||\Phi||)$ are \emph{unknown} finite scalars (owing to Assumption 1). 

The control law is designed as
\begin{subequations}
\begin{align}
    \tau &=  -\Lambda s -  \rho (s/ ||s||)   \label{input}\\ 
    \rho &= \hat{K}_{0} + \hat{K}_{1} ||\xi|| +  \hat{K}_{2} ||\xi||^2,
\label{rho}
\end{align}
\label{control_law}
\end{subequations}
where $\Lambda$ is a positive definite user-defined gain matrix  
and $\hat{K}_i$ ($i$ = 0, 1, 2) are the estimates of ${K}^*_i$ adapted via the
following law:
\begin{equation}
\dot{\hat{K}}_{i}(t) = ||s(t)|| ||\xi||^i - \nu_{i} \hat{K}_{i}(t), \quad \hat{K}_{i}(0) > 0
\label{adaptive_law}
\end{equation}
with $\nu_{i} \in  \mathbb R^{+} $ ($i = 0,1,2$) being user-defined scalars. Eventually, $u_1$ is computed via (\ref{con}) and applied to the system as $R_B^W$ is an invertible rotational matrix.

\subsection{Closed-loop Stability Analysis}
\begin{theorem}
Under Assumption 1 and Properties 1-3, the closed-loop trajectories of (\ref{dot_s})  with control laws (\ref{control_law}) along with the adaptive laws (\ref{adaptive_law}) are Uniformly Ultimately Bounded (UUB).
\end{theorem}
\textbf{\textit{Proof.}} The closed-loop stability analysis is carried out using the following Lyapunov function
\begin{equation}\label{lyp}
    V = \frac{1}{2} s^T {M} s + \frac{1}{2} \sum_{i=0}^{2}(\hat{K}_{i} - {K}^*_{i})^2.
\end{equation}
Using (\ref{dot_s}) and (\ref{input}), the time derivative of (\ref{lyp}) yields
\begin{align}
\dot{V} = & s^T {M} \dot{s} + \frac{1}{2}s^T \dot{{M}}s + \sum_{i=0}^{2}(\hat{K}_{i} - {K}^*_{i})\dot{\hat{K}}_{i}  \nonumber \\
= & s^T(\tau - {C}s + \varphi)  +  \frac{1}{2}s^T \dot{{M}}s + \sum_{i=0}^{2}(\hat{K}_{i} - {K}^*_{i})\dot{\hat{K}}_{i} \nonumber\\
= & s^T(\tau  + \varphi) + \frac{1}{2}s^T(\dot{{M}} - 2{C})s  + \sum_{i=0}^{2}(\hat{K}_{i} - {K}^*_{i})\dot{\hat{K}}_{i} \nonumber \\ 
=& s^T (-\Lambda s - \rho (s/||s||) + \varphi)+ \sum_{i=0}^{2}(\hat{K}_{i} - {K}^*_{i})\dot{\hat{K}}_{i},
\label{lyap_dot_3}
\end{align}
where the relation $s^T (\dot{{M}} - 2C)s = 0$ from Property 3 is applied. 
The adaptive law (\ref{adaptive_law}) yields
\begin{equation}
(\hat{K}_{i} - {K}^*_{i}) \dot{\hat{K}}_{i} = ||s||(\hat{K}_{i} - {K}^*_{i})||\xi||^i + \nu_{i} \hat{K}_{i} {K}^*_{i} -\nu_{i} \hat{K}_{i}^2. 
\label{lyap_dot_4}
\end{equation}
Using (\ref{lyap_dot_3}), (\ref{lyap_dot_4}) and the upper bound structure of $||\varphi ||$ as in (\ref{up_bound}), we have 
\begin{align}
\dot{V} \leq & -s^T \Lambda s - \rho ||s|| + ||\varphi || ||s||+ \sum_{i=0}^{2}(\hat{K}_{i} - {K}^*_{i})\dot{\hat{K}}_{i} \nonumber \\
\leq & -s^T \Lambda s - \sum_{i=0}^{2}(\hat{K}_{i} - {K}^*_{i})(||s|| ||\xi||^i - \dot{\hat{K}}_{i} )  \nonumber \\
\leq & - \lambda_{\min}(\Lambda)||s||^2 + \sum_{i=0}^{2}(\nu_{i} \hat{K}_{i} {K}^*_{i} -\nu_{i} \hat{K}_{i}^2 ) \nonumber \\
\leq & - \lambda_{\min}(\Lambda)||s||^2 - \sum_{i=0}^{2} \frac{\nu_{i}}{2}  ((\hat{K}_{i} - {K}^*_{i})^2 - {{K}^*_{i}}^2 ). \label{new2}
\end{align}

Further, the definition of Lyapunov function yields
\begin{equation}
V \leq \frac{1}{2} \overline{m}||s||^2 + \frac{1}{2} \sum_{i=0}^{2}(\hat{K}_{i} - {K}^*_{i})^2. \label{new}
\end{equation}
Substituting (\ref{new}) into (\ref{new2}), $\dot{V}$ is simplified to
\begin{equation}
\dot{V} \leq - \varrho V + \zeta,  \label{new3}
\end{equation}
where $\varrho \triangleq  \frac{\min( \Lambda,{\nu_{i}}/{2} )} {(\overline{m}/2), (1/2)} >0$ and $\zeta = \frac{1}{2} \sum_{i=0}^{2}(\nu_{i} {{K}^*_{i}}^2)$. Defining a scalar $\kappa$ such that $0<\kappa<\varrho $, $\dot{V}$ in (\ref{new3}) yields
\begin{align}
\dot{V} &= -\varrho V - (\varrho - \kappa)V + \zeta.
\end{align}

Defining a scalar $\mathcal{  B} = \frac{\zeta}{(\varrho - \kappa)}$, it can be noticed that $\dot{V} (t) < - \kappa V (t)$ when $V (t) \geq \mathcal{ B}$, so that
\begin{align}
    V & \leq \max \{ V(0), \mathcal{ \bar B} \}, \forall t \geq 0,
\end{align}
and the closed-loop system remains UUB (cf. UUB definition $4.6$ as in \cite{khalil2015nonlinear}).
\begin{remark}
For continuity in control law, the term $s/||s||$ in (\ref{input}) is usually replaced by a smooth function $\frac{s}{\sqrt{||s||^2+\delta}}$ with $\delta$ being a positive user-defined scalar \cite{roy2021adaptive}. This does not alter the overall UUB stability result, and hence, repetition is avoided.
\end{remark}

\section{Experimental Scenarios, Results and Analysis}\label{sec:experiments}
In this section, we conduct experiments using the developed passive gripper and the adaptive control law to demonstrate the effectiveness of the integrative approach.

\subsection{Experimental Setup}
 \begin{figure*}
    \includegraphics[width=\textwidth]{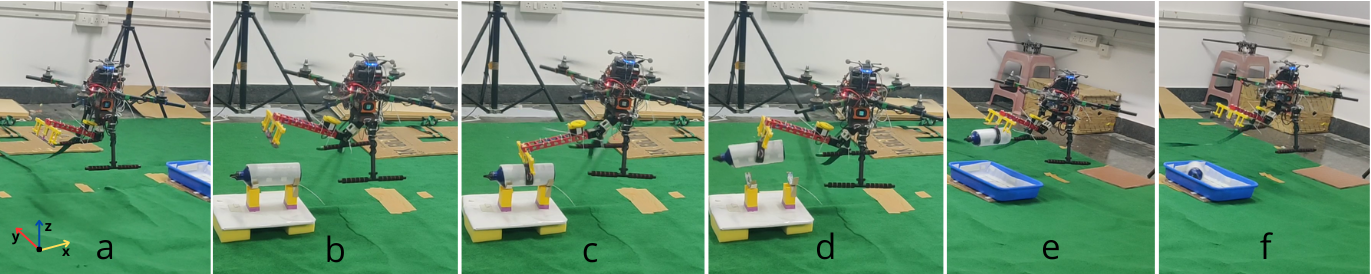}
    \centering
    \caption{\textbf{Scenario 1:} Sequence of operations of the UAM during the experiment with the proposed controller: (a) takeoff from the ground; (b) follows the trajectory and before pickup; (c) arm movement during the payload pickup (d) stabilizing itself after the payload pickup; (e) moving to the drop location; (f) dropping the payload at a predefined tray. The quadrotor is tied to the floor using a rope for safety reasons. }
    \label{fig:exp_snap_1}
\end{figure*}

The UAM system used in the experiment includes an S-650 quadrotor system (uses SunnySky V4006 brushless motors) and a 2R serial-link manipulator system (uses Dynamixel XM430-W210-T motors). Our customized bistable gripper is attached as an end-effector to the manipulator for payload pick-and-place operation. The overall setup weighs $\sim$3.0 kg. A U2D2 Power Hub Board is used to power the manipulator and the gripper. Raspberry Pi-4 is used as a processing unit that uses a U2D2 communication converter. Optitrack motion capture system (at 120fps) is used to obtain the system pose and, further, IMU data is fused to obtain necessary state-derivatives.

To verify the effectiveness of the integrated approach, the UAM was tasked to pick up a payload ($\sim0.2$ kg) from a location that was fixed and was made known to the quadrotor using the motion capture markers. The developed passive gripper and adaptive controller are evaluated with the above-mentioned UAM in two different scenarios as described in the following subsections. The following control parameters are used for the proposed controller during all the experiments:
$\Phi = \diag \{1.0, 1.0, 1.2, 1.1, 1.1, 1.0, 1.2, 1.2 \}$, 
$\Lambda = \diag \{2.0, 2.0, 3.5, 1.5, 1.5, 1.2, 3.0, 3.0 \}$, 
$\hat{K}_{0}(0) =  \hat{K}_{1}(0) = \hat{K}_{2}(0) = 0.1$, 
$\nu_{0} = 2.0$; $\nu_{1} = \nu_{2} = 5.0$, $\delta = 0.1$. 

\subsection{Scenario 1: Description, Results, and Analysis} This experimental scenario consists of the following sequences to pick an object lying horizontally  (cf. Fig. \ref{fig:exp_snap_1}):

\begin{itemize}
    \item The quadrotor takes off from its origin ($x = y = z= 0$) to achieve a desired height $z_d=1$ m with the initial manipulator joint angles $\alpha_1 (0) = \ang{0}$ and $\alpha_2(0) = \ang{110}$. 
    \item The quadrotor starts moving toward its desired location $x_d = -0.8$ m, $y_d = 0$ m to pick up the payload. 
    \item Manipulator moves its arm to desired angles $\alpha_{1d} = \ang{45}$ and $\alpha_{2d} = \ang{70}$ at $t = 10$ s (approx.) before grasping and to desired angles $\alpha_{1d} = \ang{45}$ and $\alpha_{2d} = \ang{35}$ to grab the object at $t = 12$ s (approx.). Such arm movement is necessary to generate the triggering force. After grabbing the object, the arm again moves back to desired angles $\alpha_{1d} = \ang{45}$ and $\alpha_{2d} = \ang{70}$ at $t = 14$ s (approx.).
    \item The quadrotor starts moving towards the drop point ($x_d = 0.8$ m, $y_d = 0$ m) and drop payload at $t = 31$ s (approx.) from the height $z_d=1$ m.
\end{itemize}
The main innovation of the proposed controller is to tackle uncertainties without their a priori knowledge. To properly highlight its benefit, the performance of the proposed controller is compared with the adaptive sliding mode controller (ASMC) \cite{liang2022adaptive}, which requires precise knowledge of inertial parameters in its adaptive control scheme. For parity, the sliding variable of ASMC is selected as $s$ in (\ref{slidingVar}). The other various control variables of ASMC are selected as per \cite{liang2022adaptive} after accounting for the inertial parameters of the current setup as required in \cite{liang2022adaptive}.

\begin{figure}[!h]
    \includegraphics[width=0.5\textwidth]{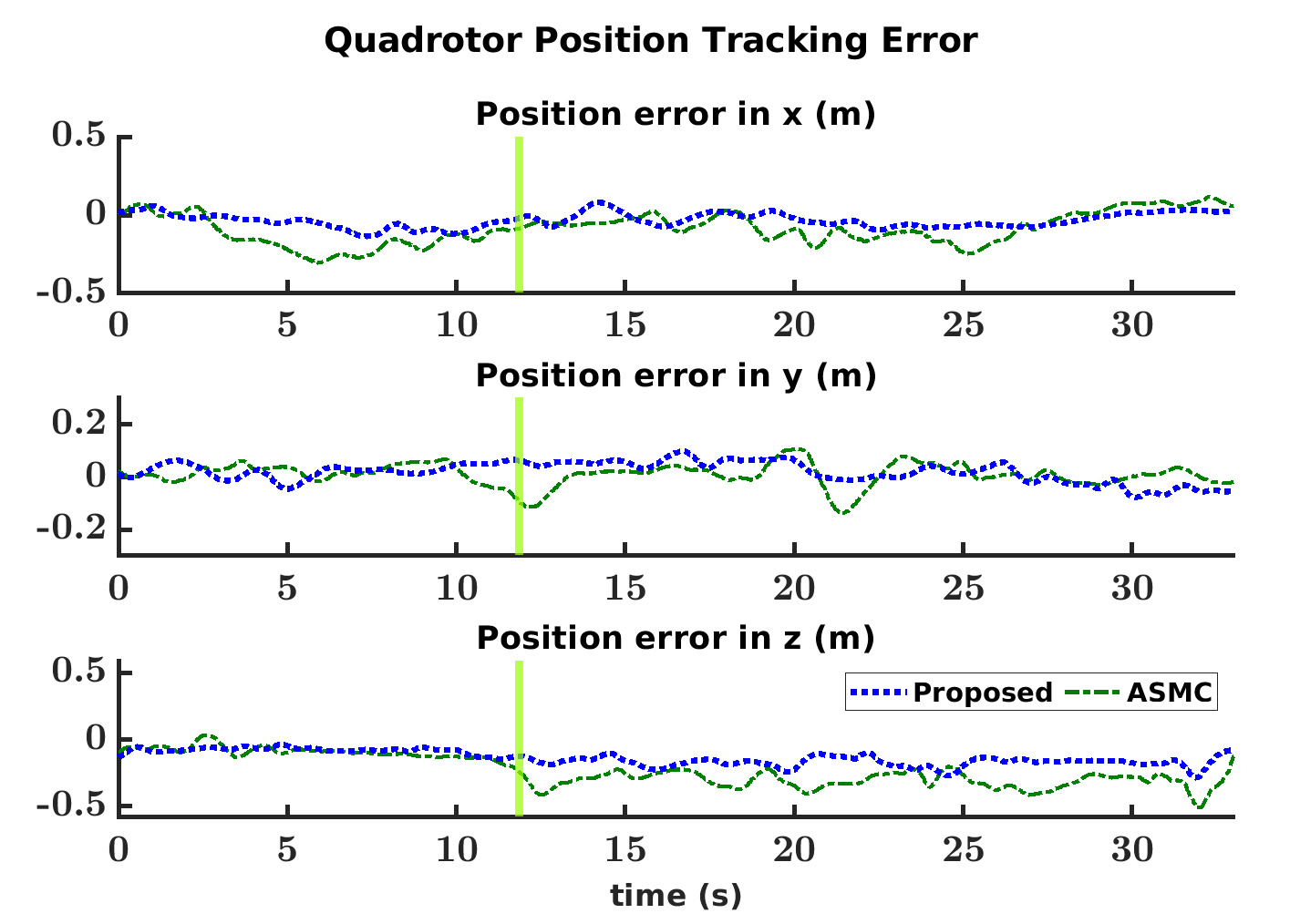}
    \centering
    \caption{Quadrotor position tracking error comparison in Scenario 1.}
    \label{plot1}
\end{figure}

\begin{figure}[!h]
    \includegraphics[width=0.5\textwidth]{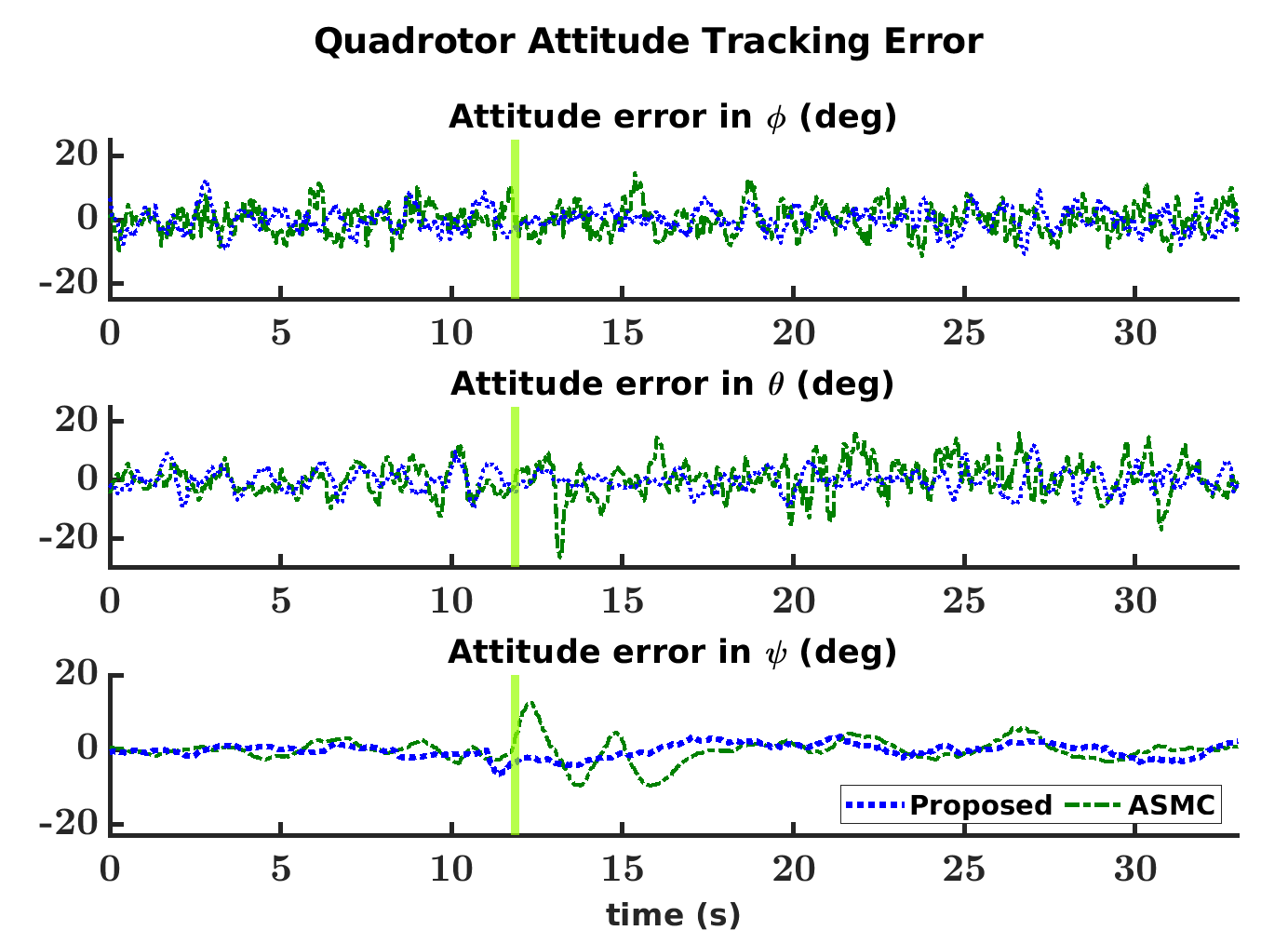}
    \centering
    \caption{Quadrotor attitude tracking error comparison in Scenario 1.}
    \label{plot2}
\end{figure}

\begin{figure}[!h]
    \includegraphics[width=0.5\textwidth]{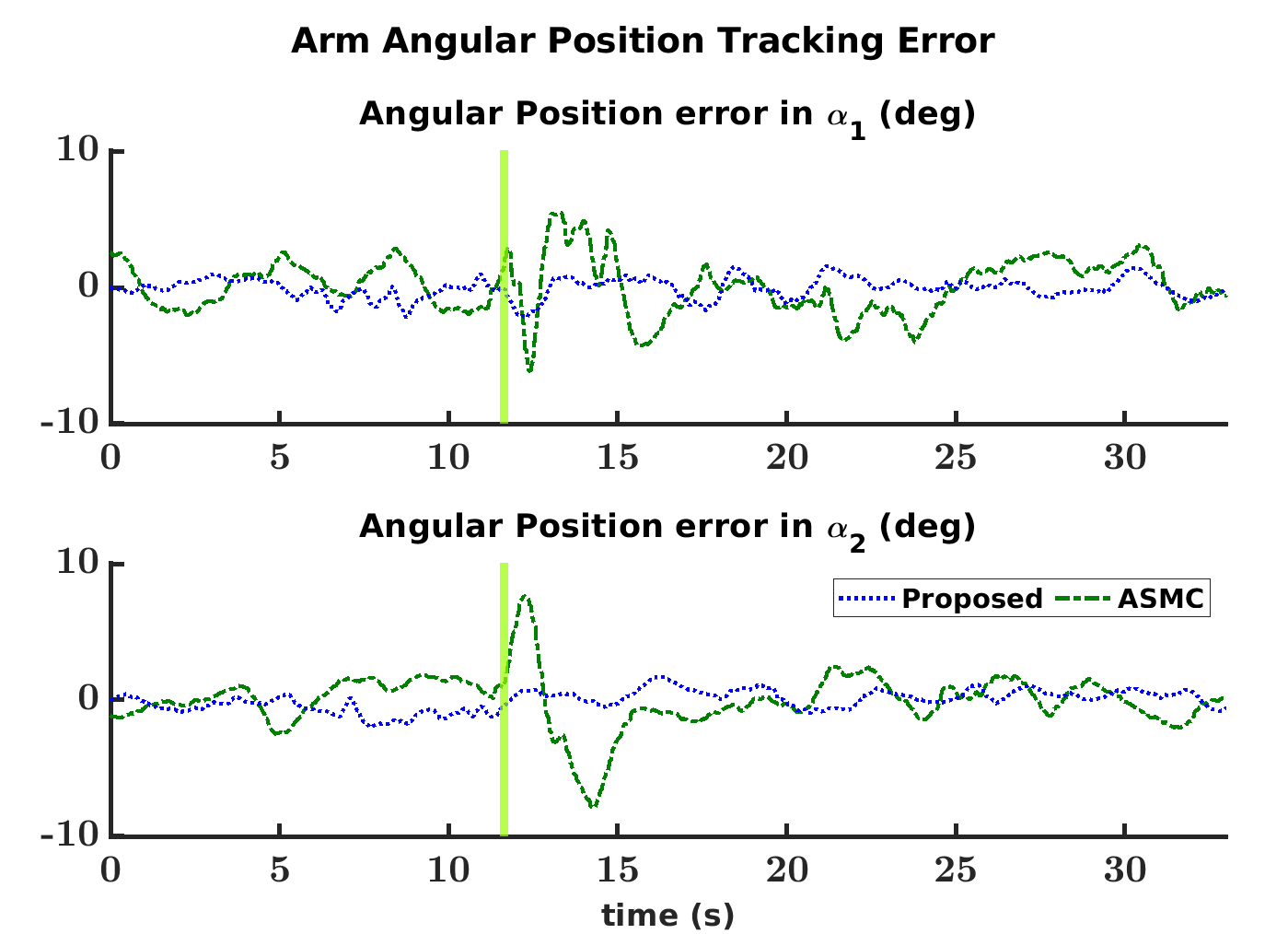}
    \centering
    \caption{Arm angular position tracking error comparison in Scenario 1.}
    \label{plot3}
\end{figure}

The performances of the controllers are highlighted via Figs. \ref{plot1}-\ref{plot3} in terms of Root-Mean-Squared (RMS) error and are further tabulated in Table \ref{table:error_p}.
The green vertical lines at $t = 12$ s in the error plots in Figs. \ref{plot1}-\ref{plot3} indicate the time when the object is picked up by the gripper. 
Abrupt spikes in the ASMC position error plots in Fig. \ref{plot3} show a significant loss in control performance right after the payload pick up at $t = 12$ s; consequently, altitude error also increases (cf. $z$ position error in Fig. \ref{plot1}). This happens because the ASMC \cite{liang2022adaptive} is not designed to tackle uncertain (state-dependent) inertial forces, impact forces, and coupling forces that occur during the payload pickup process. 
However, the proposed controller could tackle these uncertainties leading to almost negligible spikes in error plots. Further, the $\%$ error reduction data in Table \ref{table:error_p} also demonstrate the overall performance improvement of the proposed controller over ASMC.
We also note that the RMS error for the $x$ and $y$ positions using the proposed controller is less than $0.08$ m, which corresponds to the functional range of the passive gripper, while the ASMC may not work with the gripper due to a large RMS error in $x$ position. 
Here we do not consider the $z$ position since it can be compensated using the 2-DoF manipulator.

\begin{table}[!t] 
\renewcommand{\arraystretch}{1.1}
\caption{{Tracking performance comparison in Scenario 1}}
\label{table:error_p}
		\centering
{
{	\begin{tabular}{c c c c c c c c c}
		\hline
		\hline
		 & \multicolumn{3}{c}{quadrotor position} & \multicolumn{3}{c}{ quadrotor attitude} & \multicolumn{2}{c}{ arm position  }\\
    & \multicolumn{3}{c}{RMS error (m)} & \multicolumn{3}{c}{ RMS error (deg) } & \multicolumn{2}{c}{ RMS error (deg)  }\\
    \hline
		  & $x$ & $y$  & $z$ & ${\phi}$ & ${\theta}$  & ${\psi}$ & ${\alpha_1}$ & ${\alpha_2}$\\
		 \hline
		ASMC & 0.16 & 0.07  & 0.25 &  5.89 & 5.86  & 4.26 & 2.35 & 2.24  \\
		\hline 
		Proposed & 0.07 & 0.05  & 0.13 & 3.21 & 2.81  & 2.91  & 1.19 & 1.16 \\
     \hline 
		{\small $\%$ error} & {\raisebox{-1.0ex}{56.2}} &  {\raisebox{-1.0ex}{28.5}} & {\raisebox{-1.0ex}{48.0}} & {\raisebox{-1.0ex}{45.5}}  & {\raisebox{-1.0ex}{52.0}}  & {\raisebox{-1.0ex}{31.6}}  & {\raisebox{-1.0ex}{49.3}} & {\raisebox{-1.0ex}{48.2}} \\
  {\raisebox{0.5ex}{reduction}} &  &   &  &   &   &   &  &  \\
		 \hline
		\hline
\end{tabular}}}
\end{table}

\begin{figure}[!h]
    \includegraphics[width=0.49\textwidth]{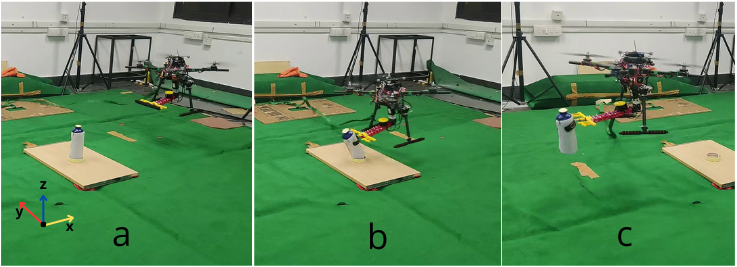}
    \centering
    \caption{\textbf{Scenario 2:} Sequence of operations of the quadrotor during the experiment with the proposed controller: (a) taking off from the ground;  (b) grasping the object  (c) stabilizing itself after picking up the object.}
    \label{fig:exp_snap_3}
\end{figure}
\subsection{Scenario 2: Description, Results, and Analysis} Scenario 1 established the effectiveness of the proposed control laws over the state-of-the-art. 
Now, it is also important to verify the swiftness of the proposed integrative approach in grasping objects. 
In this direction, a new experimental Scenario 2 is created, where a vertically placed object is to be grasped without toppling it. 
The key difference between Scenario 1 and 2 is that the first one maintains the quadrotor's posture but moves the manipulator for grasping, while the second one maintains the manipulator's posture but grasps the object via the approach speed of the quadrotor.
The experimental scenario consists of the following sequences to grasp a vertically standing object (cf. Fig. \ref{fig:exp_snap_3}): 

\begin{itemize}
    \item The quadrotor takes off from the world coordinate $x = 0.5$ m, $y = 0.0$ m to achieve a desired height of $z_d = 0.65$ m with the initial manipulator joint angles $\alpha_1 (0) = \ang{0}$ and $\alpha_2(0) = \ang{90}$. 
    \item The quadrotor starts descending toward the object at $x_d = y_d = 0$ m, $z = 0.15$ m at a constant speed. 
    \item The quadrotor grabs the object with the passive gripper (the quadrotor's approach speed generates the triggering force) at the origin and ascends to a desired location $x_d = -0.5$ m, $y_d = 0$ m, $z_d = 0.65$ m. 
\end{itemize}
We have selected three different quadrotor approach velocities for this scenario (the lowest speed can ensure a sufficient triggering force for the gripper): (i) Case 1: $\dot{x}_d = -0.4$~m/sec, $\dot{y}_d = 0$~m/sec, $\dot{z}_d = -0.4$~m/sec till $x_d = 0$ m (while descending) then $\dot{z}_d = 0.4$~m/sec for ascending.  (ii) Case 2: $\dot{x}_d = -0.3$~m/sec, $\dot{y}_d = 0$~m/sec, $\dot{z}_d = -0.3$~m/sec while descending then $\dot{z}_d = 0.3$~m/sec for ascending and (iii) Case 3: $\dot{x}_d = -0.2$~m/sec, $\dot{y}_d = 0$~m/sec, $\dot{z}_d = -0.2$~m/sec while descending then $\dot{z}_d = 0.2$~m/sec for ascending. These three different approach velocities will create three different (unknown) reaction forces between the gripper and the object, and, in effect, will test the \emph{repeatability} of the proposed adaptive controller under different operational conditions.

\begin{figure}[!h]
\includegraphics[width=0.5\textwidth]{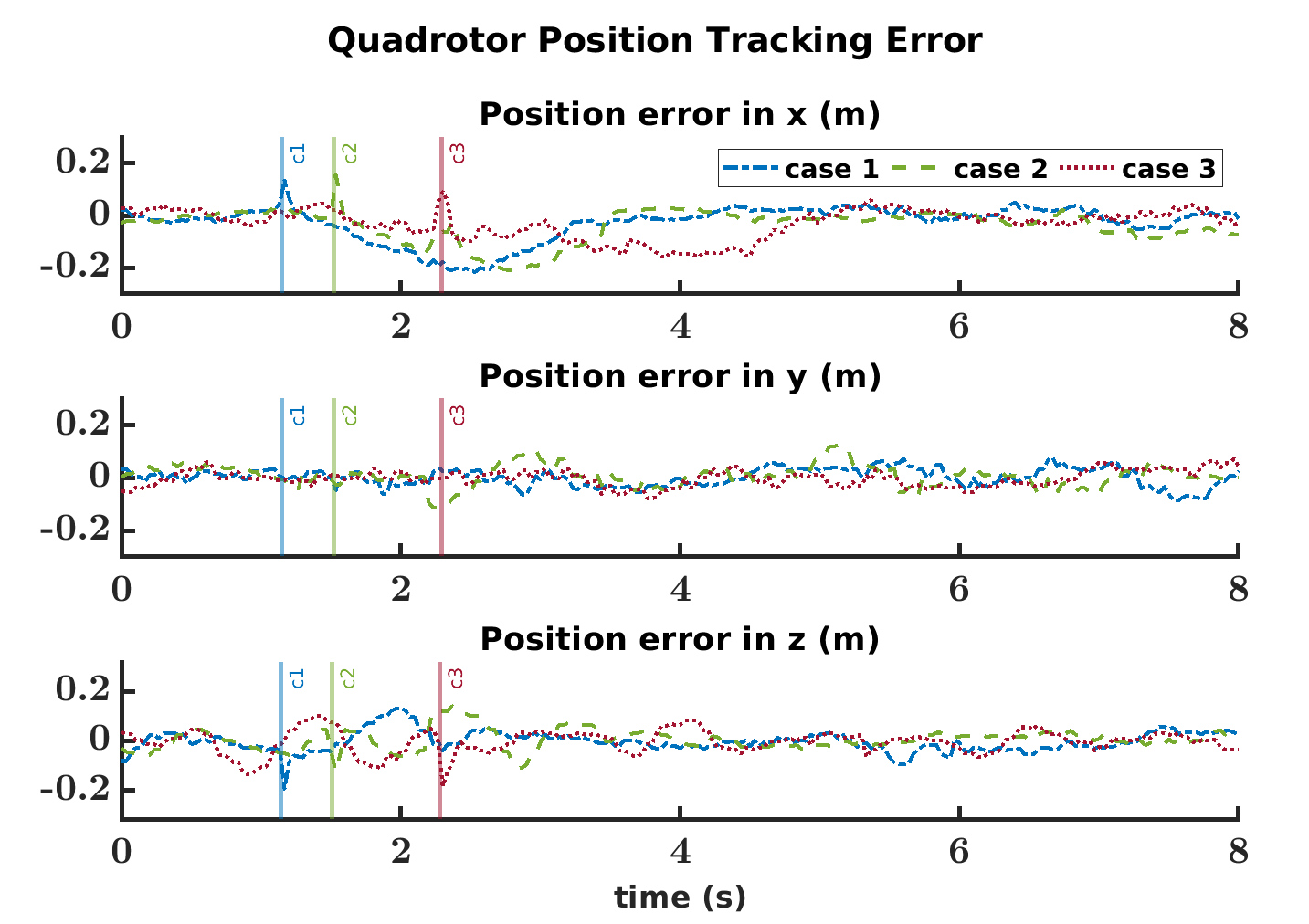}
    \centering
    \caption{Position tracking error of the proposed controller in Scenario 2 under Cases 1-3.}
    \label{plot7}
\end{figure}

\begin{figure}[!h]
\includegraphics[width=0.5\textwidth]{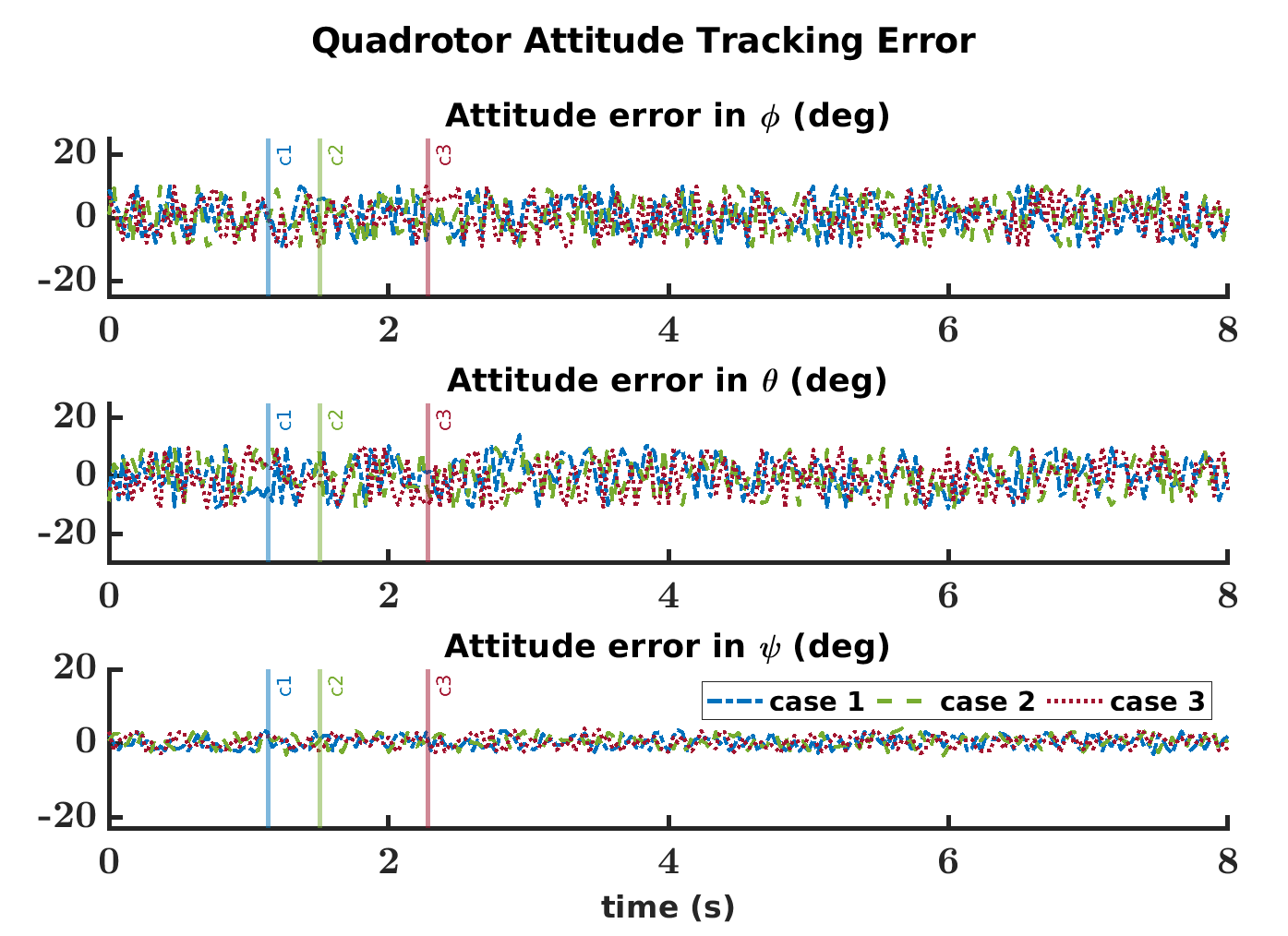}
    \centering
    \caption{Quadrotor attitude tracking error of the proposed controller in Scenario 2 under Cases 1-3.}
    \label{plot8}
\end{figure}

\begin{figure}[!h]
\includegraphics[width=0.5\textwidth]{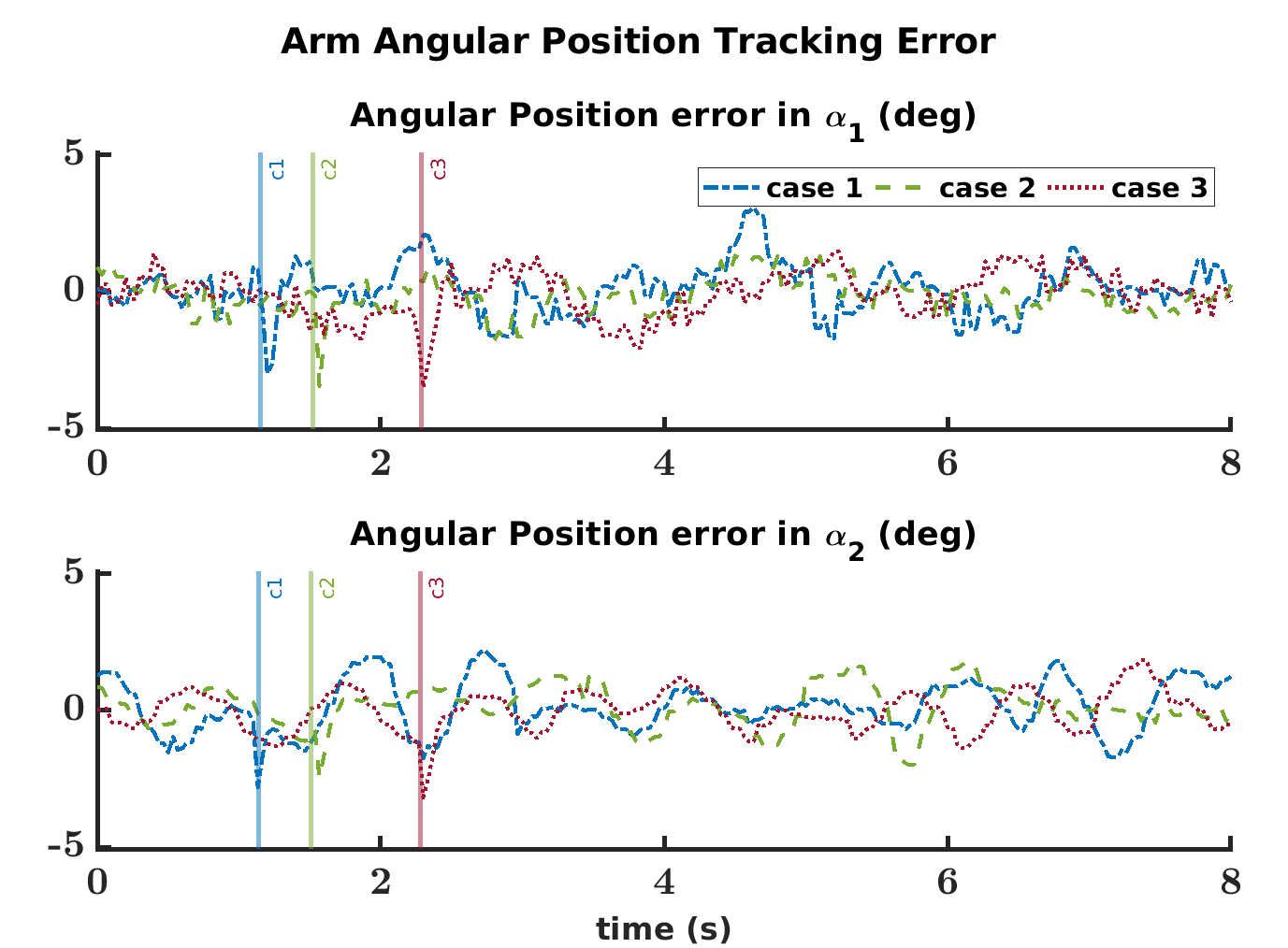}
    \centering
    \caption{Arm position tracking error of the proposed controller in Scenario 2 under Cases 1-3.}
    \label{plot9}
\end{figure}

The tracking performances of the proposed controller for Scenario 2 are depicted in Figs. \ref{plot7}-\ref{plot9} and Table \ref{table:error_p_2}. The three vertical lines in these figures ($c_1$, $c_2$, $c_3$ represent Case 1, 2, and 3 respectively) indicate the times when the object is grasped. It can be noted that the spikes appeared in the $x$, $z$ plots of the quadrotor and the arm position angles during the object grasping are successfully negotiated and damped by the controller. Consequently, the object was successfully picked up in all Cases. Importantly, the near identical RMS errors in Table \ref{table:error_p_2} demonstrate the remarkable repeatability in the control performance. 
We also note that for successful grasping in this scenario, the integrated approach for the passive gripper and the adaptive controller is indispensable. 
If the UAM is equipped with a traditional gripper, it needs to precisely control the closing of the gripper, which is a nontrivial task.

\begin{table}[!t] 
\renewcommand{\arraystretch}{1.1}
\caption{{Performance of the proposed controller in Scenario 2}}
\label{table:error_p_2}
		\centering
{
{	\begin{tabular}{c c c c c c c c c}
		\hline
		\hline
		 & \multicolumn{3}{c}{quadrotor position} & \multicolumn{3}{c}{ quadrotor attitude} & \multicolumn{2}{c}{ arm position  }\\
    & \multicolumn{3}{c}{RMS error (m)} & \multicolumn{3}{c}{ RMS error (deg) } & \multicolumn{2}{c}{ RMS error (deg)  }\\
    \hline
		  & $x$ & $y$  & $z$ & ${\phi}$ & ${\theta}$  & ${\psi}$ & ${\alpha_1}$ & ${\alpha_2}$\\
		 \hline
		Case 1 & 0.09 & 0.04  & 0.11 &  3.78 & 2.41  & 1.98 & 1.74 & 1.66  \\
		\hline 
		Case 2 & 0.08 & 0.05  & 0.13 & 3.84 & 2.61  & 1.89 & 1.53 & 1.41  \\
     \hline 
		Case 3 & 0.10 &  0.04 & 0.09 & 3.25  & 2.13  & 1.95 & 1.32 & 1.39  \\
		 \hline
		\hline
\end{tabular}}}
\end{table}

\section{Conclusions}
In this paper, we have introduced an integrated approach for aerial grasping by combining a unique bistable passive gripper and an adaptive control technique. The gripper initiates the grasping process automatically upon contact with an object, simplifying the control requirements. The proposed adaptive control strategy removes the necessity for any prior assumptions about uncertainties, whether in nominal conditions or upper limits. Through real-life experiments involving object grasping under two distinct scenarios, we have successfully demonstrated the effectiveness of the proposed integrated approach. Notably, our approach has shown marked improvements in performance compared to existing state-of-the-art methods, suggesting its potential impact on robust aerial grasping.

\bibliographystyle{IEEEtran} 
\bibliography{AerialGrasping}

\vfill

\end{document}